%% file: main.tex
\ifx \DRAFT \undefined
\documentclass[11pt,a4paper,twocolumn]{article}
\else 
\documentclass{article}
\fi 

\input{header}

\title{ 
Promising and worth-to-try future directions for \\ advancing  state-of-the-art surrogates methods\\ of agent-based models  in social and health computational sciences \\
\vspace{0.25cm}
}

\author{Atiyah Elsheikh}
\affil{
a.m.g.elsheikh@gmail.com}
\date{\today}
\date{\vspace{-3ex}}

\begin{document}

\maketitle 

\providecommand{\keywords}[1]
{
  \small	
  \textbf{\textit{Keywords---}} #1
}


\ifx \undefined \DRAFT

\else

\section*{Remarks}

The side note (s.a.\ this\actualized{Version 1.0 this is used to show recent updates and changes)} can be disabled by 
\begin{verbatim}
\def \DRAFT {} 
\end{verbatim} at the top of the TeX document. 

Please use the  comment command or suggestion command for highlighting any recommendations, suggestions, modification e.g. \suggestion{By Mike: X and Y should be followed because z and w}. 
Please place your name in the comment. 

\fi 



\begin{table*}[t]
\label{Tab:ObjectivesMethods}
\begin{center}
        \begin{tabular}{|c|l|}
       \hline
       \textbf{Objective} & \textbf{Methods} \\
       \hline 
       Accuracy  & \multirow{4}{13cm}{
       employing statistical emulators suitable for nonlinear dynamical systems  
       \begin{itemize}
           \item auto-associative model \cite{Girard2008}
           \item nonlinear Gaussian Emulators \cite{Mohammadi2019} 
       \end{itemize}} \\
        & \\
        & \\
       \hline 
       Robustness &  \multirow{3}{13cm}{instead of surrogates of single model output, \cite{Angione2022}:
       \begin{itemize}
           \item surrogates with multiple model outputs 
           \item time-series surrogates 
           emulating time-dependent trajectories of significant model state variables, e.g.\ \cite{Chen2021,Larie2021} 
       \end{itemize}} \\
        & \\
        & \\
        & \\
        & \\ 
\hline 
       Reduction of  &   \multirow{5}{13cm}{ investigating the suitability and applicability of equation-based ML techniques in the context of ABMs:
    \begin{itemize}
        \item 
        equations learning from ABM simulations \cite{Nardini2021}
        \todo{item : mean field models and coarse graining citations: Accurate and
efficient discretization for stochastic models
providing near agent-based spatial resolution at low
computational cost another citation: Nonlocal aggregation
models: a primer of swarm equilibria} 
        \item
        neural differential equations \cite{Chen2019}
        \item 
        reservoir computing techniques \cite{Lukosevicius2012} 
    \end{itemize} } \\ 
       training  & \\
       runtime & \\
       & \\
       & \\ 
\hline 
       Reliability &  \multirow{3}{13cm}{ 
       Establishing and evaluating reliability indices via
       \begin{itemize}
           \item executing model-based analysis tools upon surrogates
           \item and comparing them with application to both the actual model and the equivalent surrogate 
            \todo{See Innovation paper: Combining Calibration or SA with Surrogate constructions}
       \end{itemize}
       } \\ 
       & \\
       & \\ 
       & \\

       \hline 
    \end{tabular}
    \end{center}
    \caption{Proposed research objectives and corresponding methods}
\end{table*}

\section{Motivation}

Governments and policy makers usually need to meet sophisticated decisions for realizing complex interventions, e.g.\ regarding public health \cite{Lorenc2014}.
These decisions may largely impact the population, their daily life routines, affordability of living costs including vital basic demands, their health and well-being, among others.  
Mathematical modelling can become a useful tool for analysing the unthought impact of such complex interventions and potentially anticipating unexpected or undesired outcomes.   

One widely employed category of mathematical modelling paradigms in SHCS is agent-based modelling \cite{Silverman2021,Boyd2022}. 
ABMs are distinguished with their wide flexibility in terms of micro-level representation of population individuals, possibly associated with their behavior or attitude, variable model structures, possibly multi-clocked simulation processes and unpredictable model outcomes upon parameter variations.  

Constructing and implementing realistic demographic ABMs is sought in many applications of SHCS s.a.\ Epidemiology, socio/health-economics, social care, health inequalities among others.  
Nevertheless, modelling applications in SHCS belong to the most difficult categories in terms of validity, identifiability and predictive power. 
 They don't rely on first principle laws of Physics that make them experimentally reproducible. Moreover, large number of simplification assumptions need to be made to establish a "reasonable" model. The impact of such assumptions and quality  of data on model validity needs to be challenged. 
 
Overall, maximal exploitation of model-based mathematical analysis tools are thought s.a.:
\begin{itemize}
    \item 
    Monte-carlo based extensive simulations, e.g.\ for qualitative inference of various model states and emergent behviors \cite{An2016}
    \todo{another item: Inference of parameter distributions via Markov-chain Monte Carlo sampler Ref.: Social Computing, Behavioral-Cultural Modeling and Prediction}
    \item Calibration with available data \cite{Thiele2014} aiming at improving the predictive power of the model
    \item  Quantifying or classifying the impact of model parameters on model outputs via sensitivity analysis \cite{Saltelli2020} 
    \item 
    Quantifying the propagation of uncertainties in model parameters to model outputs via uncertainty quantification \cite{McCulloch2022}
\end{itemize}   
These tools need to be executed with  care due to the challenging nature of common ABMs exhibiting strongly nonlinear chaotic behavior of model states, correlation among parameters, jumps and tipping points in model outputs \cite{Broeke2016}. 
For instance, widely-used global sensitivity analysis methods like Morris \cite{Campolongo2007} and Sobol \cite{Saltelli2002} methods demonstrate limited reliability in the context of ABMs \cite{Broeke2016}. 

Overall, it is sought that successful deployment of some of the mentioned mathematical tools can provide useful insights into the reliability of model predictions, the impact of model simplification assumptions, the potential contribution of existing or desired missing data, the propagation of input uncertainties in model outputs and the effectiveness of tuning the mathematical model formulation.

 By making such mathematical package of tools accessible and computationally executable, a non-endless but hopefully effective model tuning process can be established. This process aims at systematic and mathematically-directed improvement of model descriptive and predictive power.
Nevertheless, the execution of model-based analysis for realistic large-scale models can be excessively long suffering from the computational demand exponentially proportional to the model size and the number of model parameters, given that the runtime of a single simulation can be huge when attempting to employ realistic population size.       

\section{Potential future directions}
\label{sec:aim}

A classical approach to overcome the huge computational demand for realizing model-based analysis of ABMs is to establish equivalent surrogates that accurately replicate not only the model behavior with tremendous speed up \cite{Kasim2020} but also their structural features and statistical properties \cite{Jones2001,Krityakierne2016,Lancaster2018}. 
Despite recent advances in ABM-equivalent surrogate models employing ML-based approaches \cite{Angione2022, Sivakumar2022}, other literature covering surrogate modeling techniques for 
\begin{itemize}
    \item ABMs particularly in biological application field  or
    \item other modelling paradigms of dynamical systems, e.g.\ (equation-based) system  models 
\end{itemize}  
demonstrate that there are further methods that may have good potential for advancing state-of-the-art surrogate modelling for ABMs in SHCS.

The main aim of this ad-hoc brief report is to highlight some of such potential methods that were adequate and computationally less demanding for nonlinear dynamical models in other modeling application areas, cf.\ Table 1.
To the author knowledge, these methods 
have been not, at least extensively, employed for ABMs within the field of SHCS, yet.
Thus, they might be, but not necessarily, useful in progressing state of the art for establishing surrogate models for ABMs in the field of SHCS. 
In Table 1, each research objective is to be tackled by some of (but not limited to) the corresponding stated methods.
Without guarantee of great success, their effectiveness is worth to examine. 
Significant advancement in the state-of-the-art methods for ABM-based surrogate shall enable model-based analysis with realistic population sizes.

\section*{Funding}

Till January 2024, Dr. Atyiah Elsheikh is a Research Software Engineer at MRC/CSO Social \& Public Health Sciences Unit, School of Health and Wellbeing, University of Glasgow. He is in the Complexity in Health programme. He is supported by the Medical Research Council (MC\_UU\_00022/1) and the Scottish Government Chief Scientist Office (SPHSU16). \\ 

For the purpose of open access, the author(s) has applied a Creative Commons Attribution (CC BY) licence to any Author Accepted Manuscript version arising from this submission.

\bibliographystyle{apalike}
\bibliography{ai,analysis,greencomputing,julia,systemmodel,abm}

\end{document}

%% file: header.tex
\usepackage[utf8]{inputenc}
\usepackage{amsmath,amsfonts,amssymb}
\usepackage{hyperref}
\usepackage{graphicx}
\usepackage{mathtools} 
\usepackage{enumitem}  
\usepackage{authblk}
\usepackage{parskip} 
\usepackage{multirow} 

\ifx \DRAFT \undefined 
\usepackage[disable]{todonotes} 
\else 
\usepackage{todonotes} 
\fi

\usepackage{xargs}    

\newcommand{\suggestion}[1]{\todo[inline,linecolor=green,backgroundcolor=green!25,bordercolor=green]{\textbf{Suggestion: }#1}}

\newcommand{\actualized}[1]{\todo[linecolor=red,backgroundcolor=gray!25,bordercolor=red]{\textbf{Actualized:} #1}}

\newcommand{\highlight}[1]{\textbf{\textcolor{blue}{#1}}}